\documentclass[10pt,twocolumn,letterpaper]{article}
\usepackage[pagenumbers]{cvpr}
\usepackage{xcolor}
\usepackage{colortbl}
\usepackage{booktabs}
\usepackage{threeparttable}
\usepackage{multirow}

\definecolor{XMattePurple}{HTML}{6B5CA5}
\definecolor{XMatteLight}{HTML}{AA9BCF}
\definecolor{XMatteTint}{HTML}{F2EFF8}
\definecolor{BaselineGray}{HTML}{AAB2BE}
\colorlet{XMatteRow}{XMattePurple!40}

\definecolor{cvprblue}{rgb}{0.21,0.49,0.74}
\usepackage[breaklinks,colorlinks,allcolors=cvprblue]{hyperref}
\usepackage{etoolbox}

\makeatletter
\patchcmd{\@maketitle}{\vskip .375in}{\vspace*{-20pt}}{}{\PackageWarning{title}{Patch failed}}
\patchcmd{\@maketitle}{\vspace*{24pt}}{\vspace*{16pt}}{}{}
\makeatother

\title{X-Matte: VAE-Free Pixel-Space Flow Matching for Single-Step Image Matting}

\author{
Zanyi Wang$^{1}$ \quad Jiaang Li$^{2}$ \quad Dengyang Jiang$^{3}$ \quad Ziru Wang$^{4}$ \quad Chengzu Li$^{5}$\\
Davit Abrahamyan$^{1}$ \quad Harry Yang$^{3}$ \quad Serge Belongie$^{2}$ \quad Mengmeng Wang$^{6}$\\[4pt]
$^{1}$UC San Diego \quad $^{2}$University of Copenhagen \quad $^{3}$HKUST\\
$^{4}$Independent \quad $^{5}$University of Cambridge \quad $^{6}$Zhejiang University of Technology
}

\begin{document}

\twocolumn[{%
\renewcommand\twocolumn[1][]{#1}%
\maketitle
\vspace{-0.3cm}
\begin{center}
\captionsetup{type=figure}
\includegraphics[width=\textwidth]{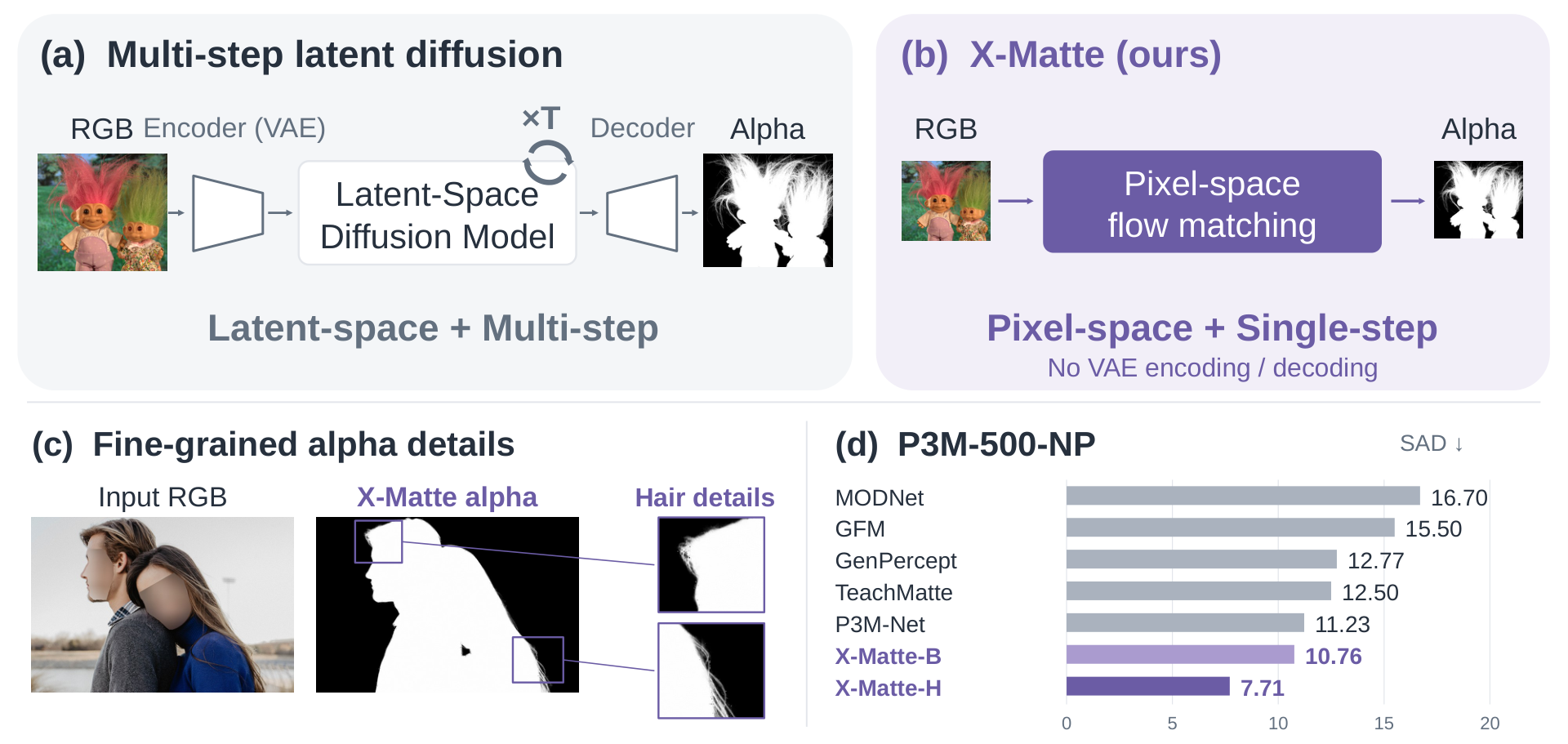}
\caption{\textbf{Overview of X-Matte.} (A) A representative multi-step latent diffusion pipeline for image matting, involving VAE encoding, decoding, and iterative inference. (B) X-Matte performs direct pixel-space alpha prediction without VAE encoding or decoding and without iterative sampling. (C) Qualitative visualization of fine alpha details. (D) SAD comparison on P3M-500-NP against representative automatic trimap-free matting methods.}
\label{fig:style}
\end{center}
}]

\begin{abstract}
Pre-trained generative models offer rich visual priors for image matting, but their adaptation often relies on variational autoencoders (VAEs), whose spatial compression can degrade fine alpha details and introduce additional inference overhead. To address these limitations, we propose X-Matte, a VAE-free pixel-space flow matching framework for single-step image matting. By eliminating latent compression, X-Matte formulates the RGB-to-alpha transition as a continuous flow directly in pixel space. We adopt trajectory-wide $x$-prediction with transition-aware alpha supervision, enabling the model to learn from intermediate states while directly predicting the target alpha matte in a single forward pass. This design avoids both VAE encoding and decoding and iterative sampling. Extensive experiments demonstrate performance competitive with state-of-the-art trimap-free matting methods on P3M benchmarks, together with promising zero-shot generalization to RWP-636 and AIM-500. Moreover, X-Matte achieves up to a $9.4\times$ inference speedup over a constructed single-step latent counterpart at high resolution, demonstrating the effectiveness of VAE-free pixel-space modeling for accurate and efficient image matting.
\end{abstract}

\section{Introduction}

Image matting estimates the continuous per-pixel opacity of foreground objects, known as the alpha matte \cite{levin2007closed, cai2019disentangled, levin2008spectral}. Unlike binary segmentation, matting requires accurate recovery of fine structures and semi-transparent boundaries, such as hair and fur, where small prediction errors can produce visible artifacts in image compositing \cite{porter1984compositing, chuang2001bayesian}. Recent discriminative methods, ranging from specialized CNNs to Vision Transformers (ViTs), commonly formulate matting as dense pixel-wise regression \cite{yao2024vitmatte, ke2022modnet, dosovitskiy2020image}. Despite their strong performance, generalization to unseen foreground objects and imaging conditions remains challenging \cite{li2021privacy, yao2024matte}. Pre-trained generative models offer a promising alternative by providing rich visual priors learned from large-scale image data. The key challenge is therefore how to adapt these priors to pixel-accurate alpha estimation without sacrificing fine structures or inference efficiency.

Existing generative matting methods, many of which build on Latent Diffusion Models (LDMs) \cite{rombach2022high}, largely fall into two paradigms. \textbf{Iterative latent generation} encodes alpha mattes into the latent space of a Variational Autoencoder (VAE) \cite{ge2025generative, wang2024matting} and commonly recovers them through iterative noise-to-alpha sampling \cite{song2020denoising, wang2024matting}. 
However, such VAEs are designed for natural RGB images rather than pixel-accurate alpha estimation, and their compression can discard spatial information to which fine structures and semi-transparent boundaries are particularly sensitive, while iterative sampling adds substantial overhead to what is ultimately a deterministic prediction. Fixed-timestep regression instead applies a pre-trained diffusion backbone at a fixed timestep for efficient dense prediction \cite{xu2024matters, xiang2025teaching, tang2023emergent, zhu2024exploring}, often with task-specific decoders that recover fine details from latent features \cite{xu2024matters, xiang2025teaching}. However, this paradigm still relies on compressed representations, incurs VAE overhead at high resolutions, and does not explicitly exploit supervision along a task-specific RGB-to-alpha trajectory. Both paradigms further face a resolution gap, as generative backbones are typically pre-trained at fixed resolutions far below those required for high-fidelity matting.

These limitations raise a natural question: \textbf{Can a pre-trained generative model predict alpha mattes directly in pixel space, without a VAE interface or iterative sampling?} 
We address this question with \textbf{X-Matte}, which builds on a pre-trained pixel-space generative backbone and formulates the RGB-to-alpha transition as a continuous flow, as illustrated in Figure~\ref{fig:style}.
 Rather than encoding alpha into an RGB-oriented latent space or regressing from a single fixed state, X-Matte learns across the RGB-to-alpha trajectory and operates directly at native image resolution. This provides a direct path from generative priors to alpha estimation, avoiding VAE compression and reconstruction while retaining single-step inference.

Specifically, X-Matte addresses the above limitations in three ways. First, we directly adapt the pre-trained pixel-space backbone for RGB-to-alpha prediction, removing the dependence on compressed representations and the associated encoding and decoding overhead. Second, we adopt $x$-prediction to estimate the target alpha at intermediate timesteps, enabling trajectory-wide training with transition-aware alpha supervision. At inference, the original RGB image serves as the starting state, allowing the model to predict the alpha matte in a single forward pass. Thus, unlike fixed-timestep regression, single-step inference does not restrict training to a single state (Table~\ref{tab:paradigm}).  Third, we adapt positional embeddings during fine-tuning and interpolate them for different inference resolutions, bridging the gap between generative pre-training and high-resolution matting.

Experiments on P3M benchmarks demonstrate competitive matting accuracy, while zero-shot evaluations on RWP-636 and AIM-500 show cross-domain generalization. Our main contributions are summarized as follows:

\begin{itemize}
    \item \textbf{VAE-Free Pixel-Space Matting.}
    We introduce X-Matte, a generative matting framework that directly adapts pre-trained pixel-space models for RGB-to-alpha prediction, eliminating VAE compression and reconstruction.

    \item \textbf{Trajectory-Wide Alpha Prediction.}
    We formulate a task-specific RGB-to-alpha flow with $x$-prediction and transition-aware supervision, enabling training across intermediate states and single-step inference.

    \item \textbf{Accurate and Efficient Matting.}
    X-Matte achieves competitive accuracy on P3M benchmarks, demonstrates zero-shot generalization to RWP-636 and AIM-500, and delivers up to a $9.4\times$ inference speedup over a constructed single-step latent counterpart.
\end{itemize}

\begin{figure*}[t]
    \centering
    \includegraphics[width=0.92\textwidth]{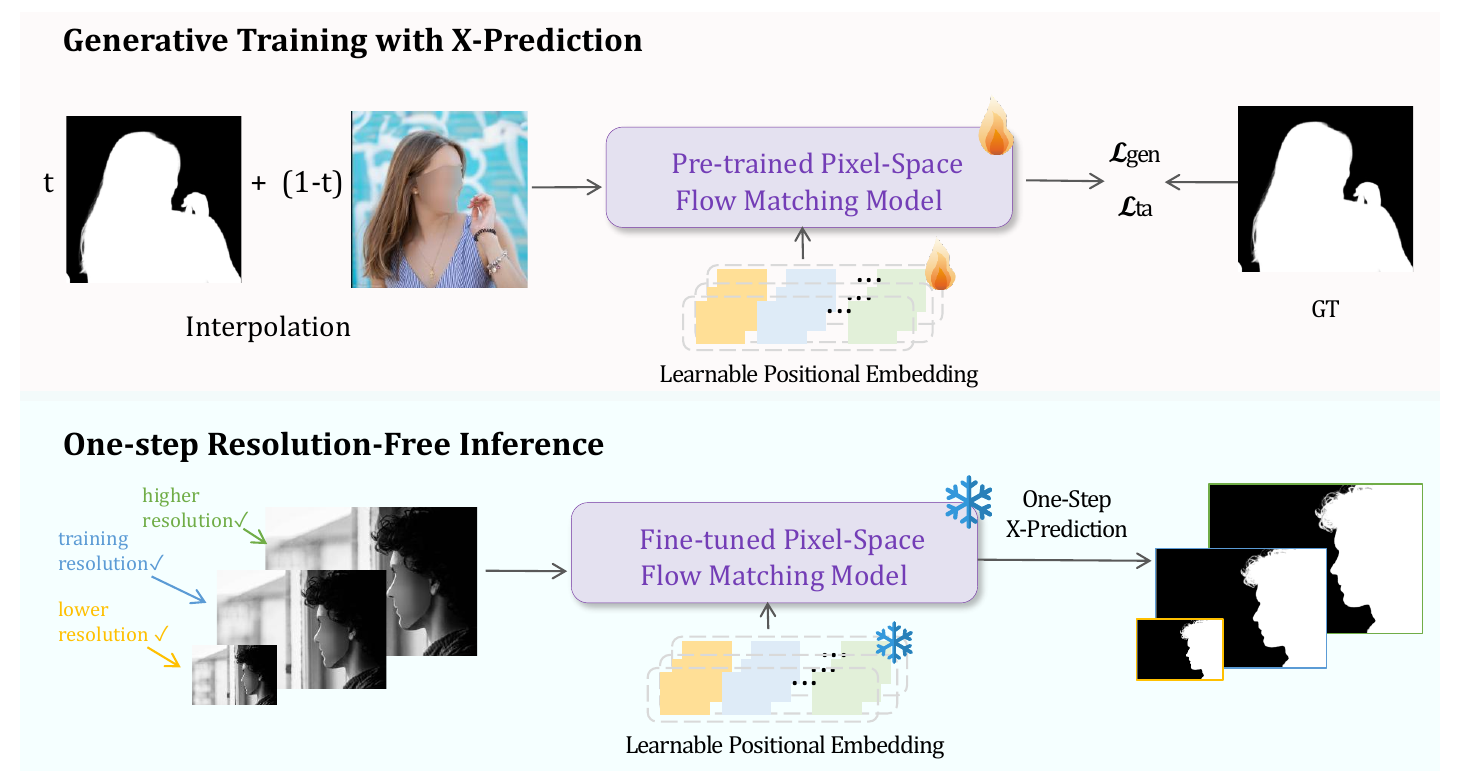}
    \caption{\textbf{Overview of X-Matte.} 
        \textbf{Top:} During training, the interpolated state 
        $\mathbf{z}_t = t\alpha_{3c} + (1-t)\mathbf{x}$ is supervised 
        by $\mathcal{L}_{\mathrm{gen}}$ and $\mathcal{L}_{\mathrm{ta}}$ 
        at every timestep via $x$-prediction—directly in pixel space, 
        without VAE compression. The learnable positional embedding 
        (flame icon) adapts to matting resolution during fine-tuning. 
        \textbf{Bottom:} At inference, a single-step $x$-prediction 
        maps RGB directly to alpha, with dynamic positional embedding 
        interpolation enabling arbitrary input resolutions.}
    \label{fig:main}
\end{figure*}
\section{Related Work}

\subsection{Discriminative Image Matting}

Image matting has predominantly been formulated as discriminative
pixel-wise regression. Early CNN-based methods, including Deep Image
Matting~\cite{xu2017deep} and FBA Matting~\cite{forte2020f},
focus on recovering high-frequency alpha details. Trimap-free methods
such as MODNet~\cite{ke2022modnet} and Background
Matting~\cite{sengupta2020background, lin2021real} further remove
manual guidance for portrait or fixed-background settings. More
recently, Vision Transformers have been introduced to matting to
exploit global context and pre-trained visual representations
\cite{yao2024vitmatte, park2022matteformer, cai2022transmatting,
ye2024unifying}. State-of-the-art discriminative systems also employ
substantial task-specific design; for example, P3M-Net
\cite{li2021privacy} combines dual decoders with domain-specific
augmentation for privacy-preserving portrait matting.

\subsection{Generative Priors and the Latent Bottleneck}

Recent methods adapt pre-trained LDMs~\cite{rombach2022high} to
image matting, but two mismatches remain.

\textbf{Stochasticity--determinism conflict.} Natural image synthesis models a distribution, whereas supervised
matting associates each input with a deterministic target.
Multi-step diffusion methods
\cite{xu2023diffusionmat, wang2024matting, hu2024diffusion}
therefore require additional mechanisms, including trajectory
correction~\cite{xu2023diffusionmat}, customized
decoding~\cite{wang2024matting}, or iterative
refinement~\cite{hu2024diffusion}. Other approaches
\cite{xu2024matters, xiang2025teaching} fix the diffusion timestep
and use the backbone as a static feature extractor, improving
efficiency but discarding the trajectory-dependent structure studied
in generative perception
\cite{wang2025deforming, gui2025depthfm, luo2023diffusion,
wang2024exploiting}.

\textbf{Latent representation paradox.} LDM-based generative matting methods typically rely on VAEs~\cite{kingma2013auto,rombach2022high} pre-trained on natural RGB images, whose perceptually oriented latent representations are not explicitly designed to preserve continuous alpha values and fine transition boundaries. Existing approaches either encode alpha mattes into RGB-oriented latent spaces~\cite{ge2025generative,wang2024matting,huang2025sdmatte} or employ task-specific decoders to recover fine details from compressed features~\cite{xu2023diffusionmat,xu2024matters,xiang2025teaching}. To better accommodate alpha information, recent studies explore transparency-aware representations, alpha-aware VAEs, and layered RGBA generation~\cite{zhang2024transparent,wang2025alphavae,yin2026qwen}, while other work leverages diffusion priors for joint foreground and alpha prediction~\cite{li2024drip} or reads alpha mattes directly from text-to-image DiT tokens while keeping the VAE encoder~\cite{wang2026rgb}. Rather than adapting RGB-oriented latent representations for alpha modeling, X-Matte removes the VAE interface entirely and formulates the RGB-to-alpha mapping directly in pixel space, avoiding latent compression while preserving access to fine-grained spatial information for alpha prediction.

\subsection{Pixel-Space Generation and Flow Matching}
To circumvent the semantic misalignment inherent in VAE-based latent spaces, the generative frontier has recently pivoted back to native pixel-space modeling~\cite{chen2025pixelflow, wang2025native, li2026basicsletdenoisinggenerative}. Pixel-space flow matching formulates the generative process directly at the original image resolution, establishing a strict end-to-end mapping paradigm that has demonstrated remarkable success in high-fidelity RGB image synthesis~\cite{chen2025pixelflow}. 
This naturally raises a compelling question: how can pre-trained pixel-space generative priors be effectively adapted to high-fidelity image matting, while preserving fine alpha details and avoiding the limitations of VAE-based latent representations?

However, adapting a pre-trained pixel-space backbone to a
deterministic task like matting presents a fundamental challenge:
modeling continuous generative trajectories in high-dimensional
pixel space remains difficult. Recent pixel-space generative
modeling shows that $x$-prediction stabilizes raw-pixel learning by
anchoring predictions to the data manifold
~\cite{li2026basicsletdenoisinggenerative}. 
Motivated by this finding, we adopt the time-conditioned Transformer architecture of SiT~\cite{ma2024sit} with a VAE-free pixel interface, replacing latent encoding and decoding with direct pixel patch embedding and prediction. Building on pixel-space generative pre-training, we further introduce a matting-specific RGB-to-alpha flow with trajectory-wide $x$-prediction and task-aware supervision, enabling direct alpha estimation without iterative sampling (Section~\ref{sec:method}).

\section{Method}
\label{sec:method}

We present X-Matte, a VAE-free pixel-space flow matching framework for image matting, as illustrated in Figure~\ref{fig:main}. X-Matte formulates the RGB-to-alpha transition as a continuous flow in pixel space, using trajectory-wide $x$-prediction during training and direct endpoint prediction for single-step inference. 
We describe the pixel-space flow formulation (Section~\ref{sec:flow}), transition-aware supervision (Section~\ref{sec:loss}), and resolution-adaptive positional transfer (Section~\ref{sec:pos}) in detail.

\subsection{Pixel-Space Flow Matching with $x$-prediction}
\label{sec:flow}

\paragraph{Flow formulation.}
Given an RGB image $\mathbf{x} \in \mathbb{R}^{H \times W \times 3}$ and its alpha matte $\alpha \in [0,1]^{H \times W}$, we replicate $\alpha$ to three channels ($\alpha_{3c}$) and define an RGB-to-alpha interpolation directly in pixel space:
\begin{equation}
\mathbf{z}_t = t\,\alpha_{3c} + (1-t)\,\mathbf{x}, \quad t \in [0,1],
\label{eq:interp}
\end{equation}
where $\mathbf{z}_0 = \mathbf{x}$ is the observed RGB image and $\mathbf{z}_1 = \alpha_{3c}$ is the target matte. Rather than explicitly predicting the velocity, the time-conditioned network $f_\theta$ is trained to estimate the alpha endpoint at each sampled timestep:
\begin{equation}
\mathcal{L}_{\mathrm{gen}} = \mathbb{E}_{t \sim \mathcal{U}(0,1)} \left\| f_\theta(\mathbf{z}_t, t) - \alpha_{3c} \right\|_1.
\label{eq:lgen}
\end{equation}
Unlike conventional noise-to-alpha generation, the observed RGB image serves as the source of the interpolation rather than an auxiliary conditioning signal. Sampling $t$ across the interpolation path exposes the model to intermediate RGB-to-alpha states, providing trajectory-wide endpoint supervision instead of restricting training to the RGB starting state.

\paragraph{Backbone and pixel interface.}
We adopt the time-conditioned Transformer architecture of SiT~\cite{ma2024sit}, equipped with a convolutional pixel patch embedding and a linear prediction head. Unlike the original latent-space SiT pipeline, our backbone is pre-trained directly in pixel space on ImageNet, without VAE encoding or decoding. The same pixel-space interface is retained during matting fine-tuning, allowing the pre-trained generative weights to be transferred to RGB-to-alpha prediction. Since the flow starts from the observed RGB image, X-Matte can leverage the visual priors learned during pixel-space pre-training without relying on latent representations.

\paragraph{Why $x$-prediction.}
Recent pixel-space generative modeling has demonstrated the effectiveness of direct clean-data prediction~\cite{li2026basicsletdenoisinggenerative}. For image matting, we adopt $x$-prediction to directly estimate the alpha endpoint $\hat{\boldsymbol{\alpha}}_t=f_\theta(\mathbf{z}_t,t)$ at each sampled timestep. This provides an interpretable prediction in the alpha domain, allowing task-specific reconstruction and boundary losses to supervise fine structures and semi-transparent regions throughout the RGB-to-alpha trajectory. While $\epsilon$- or $v$-prediction can also recover the alpha endpoint through an additional transformation, $x$-prediction offers a more direct interface for transition-aware supervision, as detailed in Section~\ref{sec:loss}.

\paragraph{Single-step endpoint prediction.}
The RGB-to-alpha interpolation in Eq.~\eqref{eq:interp} defines a linear path with constant target velocity $\mathbf{v}^*=\alpha_{3c}-\mathbf{x}$. Given the predicted endpoint $\hat{\boldsymbol{\alpha}}_t=f_\theta(\mathbf{z}_t,t)$, the corresponding velocity can be parameterized as
\begin{equation}
\hat{\mathbf{v}}_t = \frac{\hat{\boldsymbol{\alpha}}_t-\mathbf{z}_t}{1-t}, \quad 0 \leq t < 1.
\label{eq:velocity}
\end{equation}
At inference, X-Matte directly predicts the alpha endpoint from the RGB starting state, $\hat{\boldsymbol{\alpha}}=f_\theta(\mathbf{x},t{=}0)$, requiring only a single forward pass without numerical integration. Although inference uses only $t{=}0$, we view the interpolation path as structured supervision rather than a sampling trajectory, consistent with Table~\ref{tab:multistep}, where additional Euler steps do not help. Intermediate states reveal the alpha only partially, so predicting the endpoint from them forms a range of easier, related subproblems that share one network with the $t{=}0$ task; we hypothesize that this regularizes the single-step prediction. Under the same backbone, it reduces SAD from 36.87 to 22.16 relative to fixed-timestep regression (Table~\ref{tab:paradigm}).

\subsection{Transition-Aware Trajectory Supervision}
\label{sec:loss}

Direct endpoint prediction provides an interpretable alpha estimate at each sampled timestep, allowing matting-specific reconstruction and boundary losses~\cite{xu2017deep, forte2020f} to supervise the RGB-to-alpha trajectory.

\paragraph{Loss formulation.}

We combine a trajectory-wide alpha regression objective with a transition-aware auxiliary loss to jointly supervise alpha accuracy and fine boundary structures:
\begin{equation}
    \mathcal{L} = \mathcal{L}_{\mathrm{reg}} + \mathcal{L}_{\mathrm{ta}},
    \label{eq:objective}
\end{equation}
where $\mathcal{L}_{\mathrm{reg}}$ applies spatial weighting to the endpoint regression objective in Eq.~\eqref{eq:lgen}, encouraging accurate alpha prediction across the sampled RGB-to-alpha trajectory. Meanwhile, $\mathcal{L}_{\mathrm{ta}}$ combines an additional weighted alpha reconstruction term with gradient supervision that emphasizes fine boundaries~\cite{xu2017deep}. These complementary objectives encourage the model to maintain overall alpha accuracy while recovering subtle local structures, particularly around hair strands and semi-transparent boundaries.

\paragraph{Transition-aware weighting.}

To emphasize semi-transparent regions and fine boundaries, we apply an alpha-dependent spatial weighting scheme to the reconstruction terms, together with gradient supervision~\cite{xu2017deep,rhemann2009perceptually}. Most pixels belong to fully opaque foreground or transparent background regions, while sparse transition regions contain fine structures and fractional alpha values that require precise estimation. We therefore assign greater importance to intermediate alpha values while retaining supervision over opaque and transparent pixels. Applied throughout the RGB-to-alpha trajectory, this weighting encourages fine-detail recovery at different timesteps. Meanwhile, gradient supervision complements pixel-wise reconstruction by penalizing local discrepancies along object boundaries and thin hair structures. Both objectives operate directly in the alpha domain without additional inference-time processing. Table~\ref{tab:ablation} and Figure~\ref{fig:ablation_loss} demonstrate the effectiveness of this design.

\subsection{Resolution-Adaptive Positional Embeddings}
\label{sec:pos}

The backbone is pre-trained at $256\times256$, while image matting benefits from higher-resolution inputs for recovering fine spatial details~\cite{liu2021towards}. However, increasing the input resolution produces a denser token grid, creating a mismatch with the positional representations learned during pre-training. To bridge this resolution gap, we adopt learnable positional embeddings during fine-tuning at $768$px~\cite{lu2024fit,dosovitskiy2020image}. This allows the positional representations to adapt to the finer spatial layout required for alpha prediction while retaining the pre-trained pixel-space backbone. At inference, the learned embeddings are bicubically interpolated to match the input token grid, allowing the same model to process different resolutions and aspect ratios without additional fine-tuning~\cite{beyer2023flexivit}. This resolution-adaptive strategy enables flexible inference without modifying the backbone architecture or introducing additional refinement modules. The effects of positional adaptation are quantitatively evaluated in Table~\ref{tab:ablation}, while Fig.~\ref{fig:resolution} illustrates the recovery of fine boundary structures at different inference resolutions.

\begin{table*}[h]
\centering
\caption{Quantitative comparison on P3M-500-P and P3M-500-NP. All methods
are evaluated on the whole image without any user priors (trimap-free). 
$\downarrow$ indicates lower is better. Best results are in \textbf{bold}. 
Results of GenPercept and TeachMatte are obtained by running their official 
code under a unified evaluation protocol; for SMat and Alpha-Free-Matting 
we report numbers from their original papers, and entries not covered 
there are marked with `-'.}
\label{tab:main_results_p3m}
\begin{tabular}{lcccccccc}
\toprule
\multirow{2}{*}{\textbf{Method}} & \multicolumn{4}{c}{\textbf{P3M-500-P}} & \multicolumn{4}{c}{\textbf{P3M-500-NP}} \\
\cmidrule(lr){2-5} \cmidrule(lr){6-9}
 & \textbf{SAD} $\downarrow$ & \textbf{MSE} $\downarrow$ & \textbf{Grad} $\downarrow$ & \textbf{Conn} $\downarrow$
 & \textbf{SAD} $\downarrow$ & \textbf{MSE} $\downarrow$ & \textbf{Grad} $\downarrow$ & \textbf{Conn} $\downarrow$ \\
\midrule
\rowcolor{gray!20}\multicolumn{9}{l}{\textit{Discriminative}} \\
LF (CVPR 19)~\cite{zhang2019late}                 & 42.95 & 19.1 & 42.19 & 18.80 & 32.59 & 13.1 & 31.93 & 19.50 \\
HATT (CVPR 20)~\cite{qiao2020attention}               & 25.99 & 5.4  & 14.91 & 25.29 & 30.53 & 7.2  & 19.88 & 27.42 \\
P3M-Net (ACM MM 21)~\cite{li2021privacy}          & 8.73  & 2.6  & \textbf{8.22}  & 13.88 & 11.23 & 3.5  & 10.35 & 12.51 \\
MODNet (AAAI 22)~\cite{ke2022modnet}             & 13.31 & 3.8  & 16.50 & 10.88 & 16.70 & 5.1  & 15.29 & 13.81 \\
GFM (IJCV 22)~\cite{li2022bridging}                & 13.20 & 5.0  & 12.58 & 17.75 & 15.50 & 5.6  & 14.82 & 18.03 \\
SMat(CVPR 24)~\cite{ye2024unifying}               & 14.77     & 5.3    & 14.26     & 17.43     & 18.93 & 6.4  & 16.17     & 15.95     \\
Alpha-Free-Matting (AAAI 25)~\cite{liu2025training} & -     & -    & -     & -     & 18.20 & 12.6 & 15.41 & 13.66 \\
\midrule
\rowcolor{gray!20}\multicolumn{9}{l}{\textit{Generative}} \\
GenPercept (ICLR 25)~\cite{xu2024matters}     & 9.75  & 2.5  & 12.63  & 8.84  & 12.77 & 2.7 & 13.44 & 10.46 \\
TeachMatte (TMLR 25)~\cite{xiang2025teaching} & 9.53 & 3.0 & 12.12   & 8.45     & 12.50     & 2.8    & 13.68     & 10.25     \\
\midrule
\rowcolor{XMatteRow}
\textbf{X-Matte-B (Ours)}             & 8.61  & 2.3  & 10.97 & 8.01  & 10.76 & 2.8  & 9.70  & 9.26  \\
\rowcolor{XMatteRow}
\textbf{X-Matte-H (Ours)}             & \textbf{7.53}  & \textbf{2.1}  & 10.25  & \textbf{7.16}  & \textbf{7.71}  & \textbf{2.0}  & \textbf{9.52}  & \textbf{7.17}  \\
\bottomrule
\end{tabular}
\end{table*}

\section{Experiments}

\subsection{Experimental Setup}

\paragraph{Dataset.}
We train and evaluate on P3M-10k~\cite{li2021privacy}, the largest privacy-preserving portrait matting benchmark. The training set contains 9,421 high-resolution face-blurred portrait images with annotated alpha mattes. Evaluation is conducted on two test sets: P3M-500-P, comprising 500 face-blurred images following the privacy-preserving training (PPT) setting, and P3M-500-NP, comprising 500 normal celebrity portraits without face obfuscation. Evaluating on both sets measures in-distribution accuracy (P3M-500-P) and cross-domain generalization (P3M-500-NP) simultaneously. To further evaluate cross-dataset generalization, we additionally test on RWP-636~\cite{yu2021mask}, a real-world portrait matting benchmark containing 636 high-resolution images collected in diverse real-world conditions, without fine-tuning on this dataset. We further evaluate zero-shot generalization on
AIM-500~\cite{li2021deep}, a natural image matting benchmark of 500
images spanning portraits, animals, plants, and transparent objects, most of which are absent from P3M training.

\paragraph{Evaluation metrics.}

We report four standard matting evaluation metrics~\cite{rhemann2009perceptually} (lower is better): SAD and MSE assess pixel-level alpha prediction accuracy, while Grad and Conn evaluate the preservation of fine boundary details and structural connectivity, respectively.

\paragraph{Implementation details.}
We pre-train X-Matte-B (140M) and X-Matte-H ($\sim$0.9B) on ImageNet-1k at $256{\times}256$ following the $x$-prediction recipe of~\cite{li2026basicsletdenoisinggenerative}. Both are
fine-tuned at $768$px with learnable positional embeddings
(Section~\ref{sec:pos}). Full hyperparameters are in the supplementary
material.

\paragraph{Inference.}
We bicubically interpolate positional embeddings on the fly for resolution-free inference. All results use single-step endpoint prediction at native aspect ratios.

\subsection{Main Results}

\begin{figure}[h]
    \centering
    \includegraphics[width=.45\textwidth]{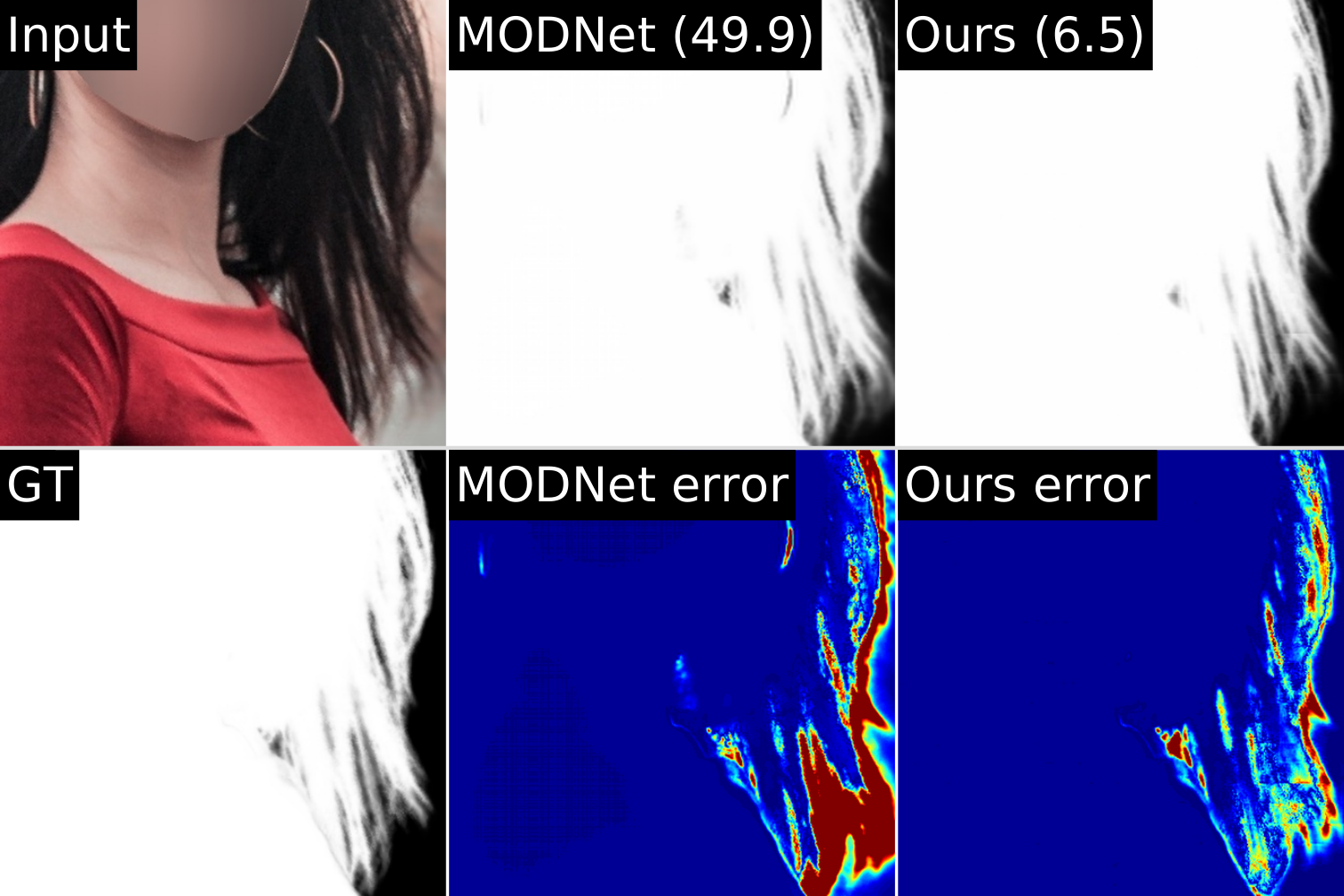} 

    \caption{\textbf{Qualitative comparison with MODNet~\cite{ke2022modnet}.} In this example, X-Matte better preserves fine hair structures and achieves lower SAD than MODNet (6.5 vs.\ 49.9).}

    \label{fig:comparison}
\end{figure}

\paragraph{Overall performance.}
As shown in Table~\ref{tab:main_results_p3m}, X-Matte-H achieves strong performance on both P3M-500-P and P3M-500-NP, obtaining SAD scores of 7.53 and 7.71, respectively. It outperforms the compared baselines on most metrics, while P3M-Net retains a lower Grad error on P3M-500-P. The smaller X-Matte-B also achieves competitive results with only 140M parameters. These results demonstrate the effectiveness of the proposed pixel-space framework for high-fidelity portrait matting. Figure~\ref{fig:comparison} further illustrates its ability to recover fine hair structures compared with MODNet.

\paragraph{Compared methods.}

We compare against representative automatic trimap-free matting methods spanning discriminative and generative paradigms. All compared methods operate on full images without user-provided trimaps, bounding boxes, or interactive guidance. Discriminative baselines include LF~\cite{zhang2019late}, HATT~\cite{qiao2020attention}, P3M-Net~\cite{li2021privacy}, MODNet~\cite{ke2022modnet}, GFM~\cite{li2022bridging}, SMat~\cite{ye2024unifying}, and the weakly-supervised Alpha-Free-Matting~\cite{liu2025training}. Generative baselines include GenPercept~\cite{xu2024matters} and TeachMatte~\cite{xiang2025teaching}, representing approaches that leverage pre-trained generative priors for dense prediction and alpha matting. We evaluate GenPercept and TeachMatte using their official code and checkpoints under a unified evaluation protocol, while results for the remaining baselines are taken from their original publications unless otherwise specified.

\begin{table}[h]
\centering
\caption{
Zero-shot generalization on RWP-636, a real-world portrait matting benchmark. All methods are trained on P3M-10k without additional fine-tuning.
}
\label{tab:rwp}
\small
\setlength{\tabcolsep}{4pt}
\begin{tabular}{lcc}
\toprule
\textbf{Method} & \textbf{SAD} $\downarrow$ & \textbf{MSE} $\downarrow$ \\
\midrule
LF (CVPR 19)~\cite{zhang2019late}          & 71.18 & 42.3 \\
HATT (CVPR 20)~\cite{qiao2020attention}        & 53.45 & 23.4 \\
MODNet (AAAI 22)~\cite{ke2022modnet}      & 42.63 & 22.7 \\
GFM (IJCV 22)~\cite{li2022bridging}         & 37.93 & 22.0 \\
SMat (CVPR 24)~\cite{ye2024unifying}        & 35.34 & 13.6 \\
\midrule
\rowcolor{XMatteRow}
\textbf{X-Matte-H (Ours)}      & \textbf{27.98}  & \textbf{12.8} \\
\bottomrule
\end{tabular}
\end{table}

\paragraph{Comparison with P3M-Net.}
Compared with P3M-Net~\cite{li2021privacy}, X-Matte-H achieves lower SAD (7.53 vs.\ 8.73), MSE (2.1 vs.\ 2.6), and Conn (7.16 vs.\ 13.88) on P3M-500-P, while P3M-Net retains a lower Grad error (8.22 vs.\ 10.25). On P3M-500-NP, X-Matte-H further improves SAD from 11.23 to 7.71 and outperforms P3M-Net across all four metrics. These results indicate that X-Matte maintains competitive boundary accuracy and performs well under the distribution shift introduced by face obfuscation.

\paragraph{Zero-shot generalization.}
We report X-Matte-H on both benchmarks. On RWP-636, X-Matte-H reaches SAD 27.98 and MSE 12.8 without any fine-tuning, achieving the best results among the compared methods (Table~\ref{tab:rwp}). AIM-500 presents a more challenging setting, as it contains diverse foreground categories beyond those commonly represented in P3M training. X-Matte-H reaches SAD 44.45, outperforming the latent-diffusion-based GenPercept at 56.36 (Table~\ref{tab:aim}). These results demonstrate promising cross-dataset generalization without task-specific fine-tuning on the target benchmarks.

\begin{table}[h]
\centering
\caption{Zero-shot generalization on AIM-500, containing
natural images entirely unseen during P3M-10k training.
All methods are trained on P3M-10k without fine-tuning.}
\label{tab:aim}
\small
\setlength{\tabcolsep}{4pt}
\begin{tabular}{lcc}
\toprule
\textbf{Method} & \textbf{SAD} $\downarrow$ & \textbf{MSE} $\downarrow$ \\
\midrule
P3M-Net (ACM MM 21)~\cite{li2021privacy}      & 103.03 & 54.5 \\
GenPercept (ICLR 25)~\cite{xu2024matters}  & 56.36  & 20.1 \\
\midrule
\rowcolor{XMatteRow}
\textbf{X-Matte-H (Ours)}         & \textbf{44.45}  & \textbf{18.8} \\
\bottomrule
\end{tabular}
\end{table}

\paragraph{Comparison with generative methods.}
Against TeachMatte~\cite{xiang2025teaching}, the closest generative
baseline, X-Matte-H improves SAD from 9.53 to 7.53 and MSE from 3.0
to 2.1 on P3M-500-P. This gap matters because TeachMatte relies on a teacher-student distillation pipeline with extra supervision signals, while X-Matte trains end-to-end with trajectory-wide
$x$-prediction and lightweight alpha-domain supervision. Against GenPercept~\cite{xu2024matters}, which adapts a latent diffusion model for dense prediction, X-Matte-H improves SAD from 12.77 to 7.71 on P3M-500-NP, while X-Matte-B achieves 10.76 with only 140M parameters. These results highlight the effectiveness of directly adapting pre-trained pixel-space generative priors to high-fidelity alpha prediction. Moreover, both GenPercept and TeachMatte rely on VAE-based latent representations, which introduce additional encoding and decoding overhead at high resolutions. In contrast, X-Matte eliminates this latent interface and directly predicts alpha mattes in pixel space, as further analyzed in Section~\ref{sec:efficiency}.

\begin{figure}[h]
    \centering
    \includegraphics[width=0.45\textwidth]{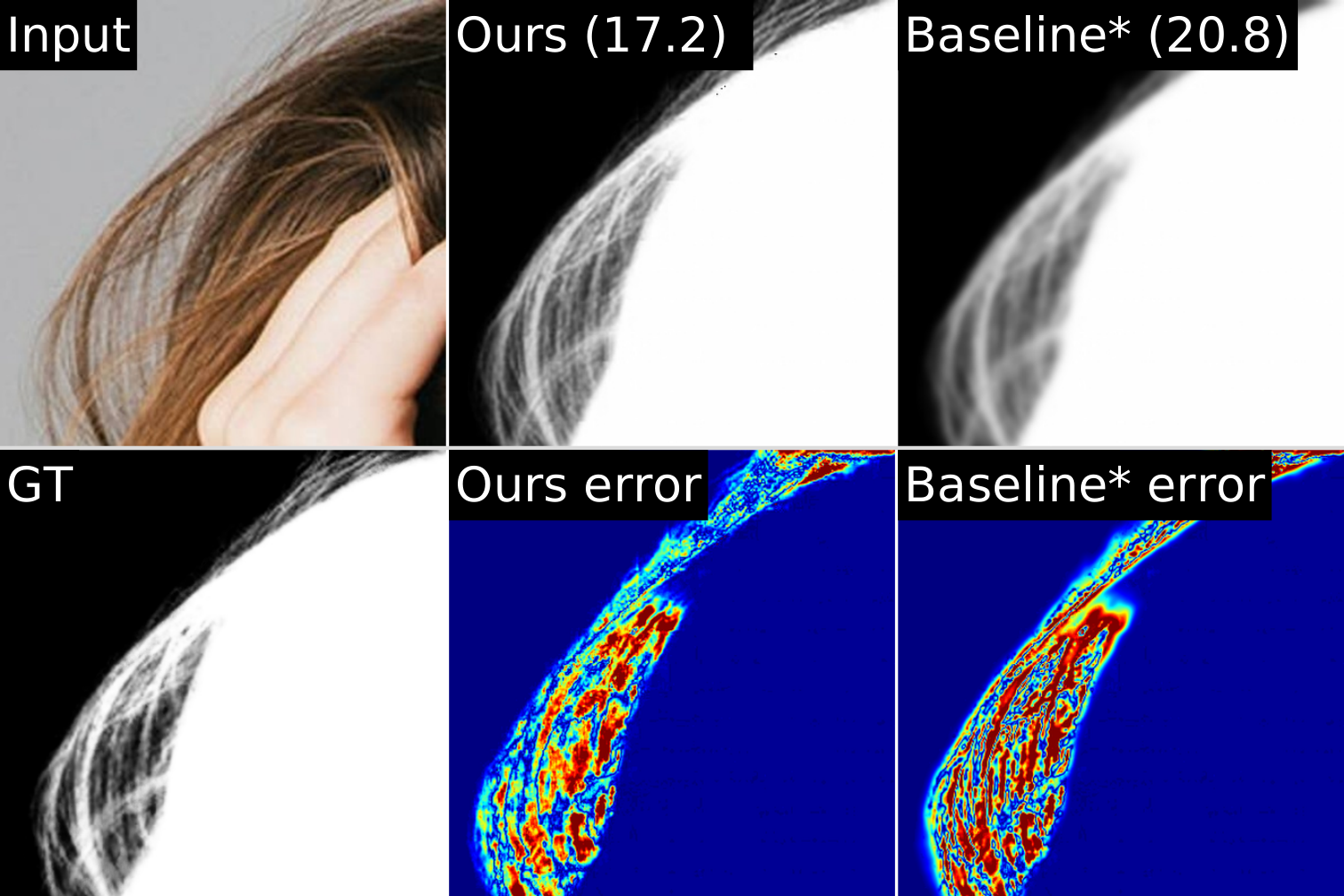}
\caption{\textbf{Effect of transition-aware weighting.}
Predicted alpha mattes and error maps comparing 
X-Matte with and without transition region weighting 
(\emph{w/o wt}). Adding the spatially adaptive weight reduces per-image SAD 
from 20.8 to 17.2, with error maps confirming 
concentrated improvement in the fine hair transition 
zone (red = high error, blue = low error).}
    \label{fig:ablation_loss}
\end{figure}

\subsection{Ablation Study}

\paragraph{Paradigm analysis.}

First, generative pre-training plays an important role in our framework. Without pre-training, both discriminative and pixel-space generative variants exhibit substantially higher errors (SAD 81.67 and 54.50, respectively). 
With ImageNet pre-training, their SAD scores decrease to 36.87 and 22.16, respectively, demonstrating the clear benefit of transferring pre-trained visual representations to image matting.

Second, trajectory-wide training substantially outperforms fixed-timestep regression under the evaluated setting (SAD 22.16 vs.\ 36.87). Both variants use the same backbone architecture and single-step inference, suggesting that supervision across intermediate RGB-to-alpha states provides an effective training strategy beyond direct regression at $t=0$.

Third, the pixel-space $x$-prediction variant achieves lower SAD than the latent-space $v$-prediction variant (22.16 vs.\ 25.17). Since this comparison changes both the representation space and the prediction target, we attribute the gain to their combination and leave disentangling the two factors to future work.

We further evaluate individual design components using the full fine-tuning recipe at $512$px, as reported in Table~\ref{tab:ablation}.

\begin{table}[h]
\centering
\caption{Paradigm ablation on P3M-500-P. All variants share 
the same backbone pre-trained on ImageNet at $256{\times}256$. 
Experiments are conducted at $256{\times}256$ to isolate 
paradigm differences under controlled conditions. The discriminative baseline fixes 
$t{=}0$ throughout training; generative variants use flow 
matching.}
\label{tab:paradigm}
\setlength{\tabcolsep}{5pt}
\begin{tabular}{lcccc}
\toprule
\textbf{Method} & \textbf{SAD}$\downarrow$ & \textbf{MSE}$\downarrow$ & \textbf{Grad}$\downarrow$ & \textbf{Conn}$\downarrow$ \\
\midrule
\multicolumn{5}{l}{\textit{Discriminative}} \\
\quad w/o pre-training    & 81.67 & 41.21 & 50.69 & 80.76 \\
\quad w/ pre-training     & 36.87 & 16.14 & 33.63 & 35.83 \\
\midrule
\multicolumn{5}{l}{\textit{Generative (latent, $v$-pred)}} \\
\quad w/o pre-training  & 60.76 & 24.01 & 37.00 & 55.95 \\
\quad w/ pre-training   & 25.17 &  7.67 & 31.46 & 21.92 \\
\midrule
\multicolumn{5}{l}{\textit{Generative (pixel, $x$-pred, ours)}} \\
\quad w/o pre-training  & 54.50 & 21.03 & 28.02 & 50.97 \\
\rowcolor{XMatteRow}
\quad \textbf{w/ pre-training} & \textbf{22.16} & \textbf{5.26} 
                                & \textbf{25.02} & \textbf{18.51} \\
\bottomrule
\end{tabular}
\end{table}

\paragraph{Component ablation.}
Table~\ref{tab:ablation} evaluates the components cumulatively. Adding $\mathcal{L}_{\mathrm{ta}}$ improves all four metrics, reducing SAD from 11.61 to 10.64, with fewer errors around hair strands and semi-transparent boundaries (Figure~\ref{fig:ablation_loss}). Pixel-space $x$-prediction makes this supervision direct, since the network outputs an alpha estimate at every timestep.

Positional embedding interpolation further reduces SAD to
9.79 without retraining, while making the embedding learnable
during fine-tuning improves it to 9.53. Finally, raising the training resolution from $512$px to $768$px reaches 8.61 SAD and improves all four metrics, validating the strategy in Section~\ref{sec:pos}.

\begin{table}[h]
\centering
\caption{Ablation study on P3M-500-P. All models use the 
same generative backbone with ImageNet pre-trained weights.
Each row cumulatively adds one component.}
\label{tab:ablation}
\begin{threeparttable}
\setlength{\tabcolsep}{3.5pt}
\begin{tabular}{lcccc}
\toprule
\textbf{Configuration} & \textbf{SAD}$\downarrow$ & \textbf{MSE}$\downarrow$ & \textbf{Grad}$\downarrow$ & \textbf{Conn}$\downarrow$ \\
\midrule
Baseline ($\ell_2$ loss)
    & 11.61 & 3.6 & 13.47 & 10.99 \\
$+\ \mathcal{L}_{\mathrm{ta}}$
    & 10.64 & 3.0 & 12.39 & 9.87 \\
$+\ $pos\_embed interp.$^{\dagger}$
    & 9.79  & 2.8 & 11.82 & 9.24 \\
$+\ $learnable pos\_embed
    & 9.53  & 2.7 & 11.54 & 9.01 \\
\rowcolor{XMatteRow}
$+\ $higher-res training
    & \textbf{8.61} & \textbf{2.3} 
    & \textbf{10.97} & \textbf{8.01} \\
\bottomrule
\end{tabular}
\begin{tablenotes}
\footnotesize
\item[$\dagger$] Bicubic interpolation of positional embeddings enables higher-resolution inference without retraining.
\end{tablenotes}
\end{threeparttable}
\vspace{-0.2cm}
\end{table}

\begin{figure}[h]
    \centering
    \includegraphics[width=0.47\textwidth]{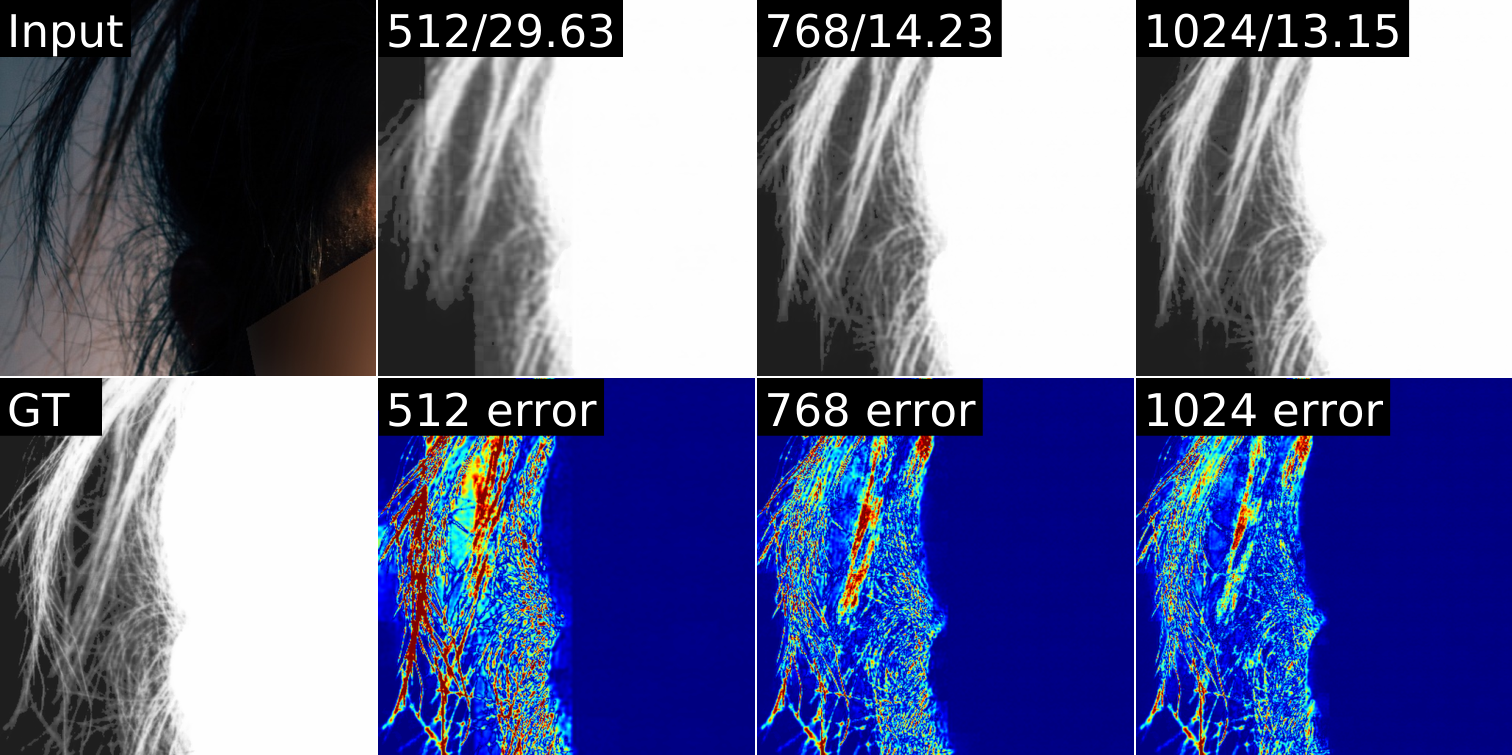}

\caption{\textbf{Effect of inference resolution.} Alpha predictions (top) and error maps (bottom) at 512px, 768px, and 1024px using the same checkpoint. The corresponding per-image SAD values are 29.63, 14.23, and 13.15. Higher resolution recovers finer hair details. Red indicates higher error and blue lower error.}
\label{fig:resolution}
\end{figure}

\paragraph{Effect of inference resolution.} We further examine the resolution adaptability of X-Matte by evaluating the same checkpoint at 512px, 768px, and 1024px without additional fine-tuning. As shown in Fig.~\ref{fig:resolution}, higher inference resolutions progressively recover finer hair strands and narrow transparent gaps. For the illustrated example, SAD decreases from 29.63 at 512px to 14.23 at 768px and 13.15 at 1024px. These observations complement the positional embedding ablation in Table~\ref{tab:ablation}, illustrating the benefit of resolution-adaptive inference for fine alpha details.

\paragraph{Number of inference steps.} We further investigate whether iterative sampling improves the performance of X-Matte. Specifically, we evaluate X-Matte-B with 1, 2, 4, and 8 inference steps on P3M-500-P using Euler integration. As shown in Table~\ref{tab:multistep}, single-step endpoint prediction achieves the lowest SAD and MSE, while additional steps provide no further improvement. This suggests that direct endpoint prediction is the preferred inference strategy under our formulation, offering both higher accuracy and lower computational cost.

\begin{table}[h]
\centering
\caption{Effect of inference steps on P3M-500-P using X-Matte-B. Lower is better.}
\label{tab:multistep}
\small
\begin{tabular}{lcc}
\toprule
\textbf{Steps} & \textbf{SAD} $\downarrow$ & \textbf{MSE} $\downarrow$ \\
\midrule
\textbf{1} & \textbf{8.61} & \textbf{2.3} \\
2 & 9.21 & 2.5 \\
4 & 9.11 & 2.5 \\
8 & 9.44 & 2.6 \\
\bottomrule
\end{tabular}
\end{table}

\subsection{Computational Efficiency of VAE-Free Design}
\label{sec:efficiency}

Table~\ref{tab:efficiency} compares X-Matte-B with our constructed VAE-based latent counterpart at two fixed input resolutions. At $512^2$ and $1024^2$, X-Matte-B reduces SAD from 15.40 to 9.84 and from 12.77 to 8.65, respectively, while reducing inference latency from 47.4\,ms to 8.7\,ms and from 177.8\,ms to 19.0\,ms, yielding $5.4\times$ and $9.4\times$ speedups. Notably, the latent counterpart also uses single-step inference, so these gains are not attributable to fewer sampling steps. Eliminating VAE encoding and decoding also reduces the total parameter count from approximately 220M to 140M. These results demonstrate that our pixel-space formulation achieves both higher matting accuracy and greater inference efficiency than the evaluated latent counterpart.

\begin{table}[h]
\centering
\caption{
Computational efficiency at high resolutions (RTX 4090, batch=1). The latent baseline is our constructed counterpart (VAE + the same SiT-B, single-step). Both are evaluated using float16 precision with memory-efficient attention (e.g., FlashAttention) for a fair comparison. SAD is evaluated on P3M-500-P under the same evaluation protocol for both ($\downarrow$ indicates lower is better).
}
\label{tab:efficiency}
\resizebox{\linewidth}{!}{%
\setlength{\tabcolsep}{4pt}
\begin{tabular}{llccc}
\toprule
\textbf{Reso} & \textbf{Method} & \textbf{Params} & \textbf{Latency} $\downarrow$ & \textbf{SAD} $\downarrow$ \\
\midrule
\multirow{2}{*}{$512^2$} 
 & Latent (VAE+SiT) & $\sim$220M & 47.4 ms & 15.40 \\

 & \textbf{X-Matte (Ours)} & \textbf{$\sim$140M} & \textbf{8.7 ms} ($5.4\times$ )& \textbf{9.84} \\
\midrule
\multirow{2}{*}{$1024^2$} 
 & Latent (VAE+SiT) & $\sim$220M & 177.8 ms & 12.77\\

 & \textbf{X-Matte (Ours)} & \textbf{$\sim$140M} & \textbf{19.0 ms} ($9.4\times$) & \textbf{8.65} \\
\bottomrule
\end{tabular}}
\end{table}

\section{Conclusion}
\label{sec:conclusion}

We presented X-Matte, a VAE-free pixel-space flow matching framework for single-step image matting. By replacing the latent diffusion interface with a direct RGB-to-alpha flow and trajectory-wide $x$-prediction, X-Matte avoids VAE compression and enables direct matting-aware supervision throughout training. It achieves accuracy competitive with state-of-the-art trimap-free matting methods on P3M benchmarks, promising zero-shot generalization to RWP-636 and AIM-500, and up to a $9.4\times$ inference speedup for X-Matte-B over a constructed single-step latent counterpart at high resolution. The improvements from X-Matte-B to X-Matte-H suggest that the framework can benefit from larger backbones. As an automatic trimap-free method, X-Matte may struggle when foreground selection is ambiguous. As illustrated in the supplementary material, the model may include additional instances of the same category or absorb adjacent dark objects and shadows into the predicted matte, particularly when foreground and background appearances are similar. These errors arise partly from foreground selection ambiguity rather than boundary reconstruction alone, due to the lack of instance-level guidance. Incorporating lightweight spatial guidance, such as points or bounding boxes, could help resolve these ambiguities more effectively in challenging scenes. Future work includes extending the RGB-to-alpha flow to video matting with temporal consistency, and to other dense prediction tasks such as depth and surface normal estimation.

\clearpage

{
\small
\bibliographystyle{plainnat}
\bibliography{main}
}

% Supplementary Material
% \clearpage
% \input{sec/X_suppl}

\end{document}